\documentclass[11pt]{article}
\usepackage{algpseudocode}
\usepackage{graphicx} %
\usepackage[final]{acl}
\usepackage{times}
\usepackage{latexsym}
\usepackage{algorithm}
\usepackage{natbib}
\usepackage{amsmath}
\usepackage{amssymb}
\usepackage{booktabs}
\usepackage{linguex}
\usepackage{utils}

\usepackage{pifont}

\usepackage[T1]{fontenc}

\usepackage[utf8]{inputenc}

\usepackage{microtype}

\usepackage{inconsolata}
\usepackage[textsize=tiny]{todonotes}

\usepackage{xurl}

\title{Which course? \textit{Dis}course!\\ 
Teaching Discourse and Generation in the Era of LLMs
}

\author{Junyi Jessy Li$^1$ \ \ Yang Janet Liu$^2$ \ \ Kanishka Misra$^1$\\\textbf{Valentina Pyatkin}$^3$ \ \ \textbf{William Sheffield}$^1$ \\
$^1$The University of Texas at Austin \\
$^2$University of Pittsburgh \\
$^3$Allen Institute for AI \\
{\small\texttt{\{jessy,kmisra,sheffieldw\}@utexas.edu, jal787@pitt.edu, valpyatkin@gmail.com}}
}

\begin{document}
\maketitle

\begin{abstract}
The field of NLP has undergone vast, continuous transformations over the past few years, sparking debates going beyond discipline boundaries. This begs important questions in education: how do we design courses that bridge sub-disciplines in this shifting landscape? This paper explores this question from the angle of discourse processing, an area  with rich linguistic insights and computational models for the intentional, attentional, and coherence structure of language. Discourse is highly relevant for open-ended or long-form text generation, yet this connection is under-explored in existing undergraduate curricula.

We present a new course, ``\textbf{Computational Discourse and Natural Language Generation}''. The course is collaboratively designed by a team with complementary expertise and was offered for the first time in Fall 2025 as an upper-level undergraduate course, cross-listed between Linguistics and Computer Science.\!\footnote{Website: \url{https://jessyli.com/courses/lin353d} \\ \indent\quad Authors listed alphabetically.} Our philosophy is to deeply integrate the theoretical and empirical aspects, and create an exploratory mindset inside the classroom and in the assignments. This paper describes the course in detail and concludes with takeaways from an independent survey as well as our vision for future directions. 
\end{abstract}

\section{Introduction}

Natural Language Processing (NLP) as a field today is one filled with intensely rapid developments and
clashing view points: 
``Scaling is powerful'' \cite{kaplan2020scaling,hoffmann2022training} vs.~``Scaling is not all you need'' \cite{li2025probing,Marcus2025,hooker2025slow}; ``LLMs cannot possibly tell us anything about language'' \cite{chomsky2023nyt,bolhuis2024three} vs.~``Actually, they can'' \cite{piantadosi2023modern, futrell2025linguistics}; ``LLMs cannot model meaning because they are not grounded'' \cite{bender-koller-2020-climbing} vs.~``But grounding is not always needed'' \cite{piantadosi2022meaning, pavlick2023symbols}; ``LLMs can be proxies for human annotators'' \cite{gilardi2023chatgpt,calderon-etal-2025-alternative} vs.~``No, that's risky'' \cite{baumann2025large,wang2025large}, to name a few. 
These debates no longer focus on NLP alone; they recontextualize how we think about linguistics and many other disciplines.
Thus, in order for the next generation of AI practitioners to grasp the cutting edge, to critically engage in these debates, and to contribute new insights, we believe it is crucial to create curricula sitting at the intersection of different (sub)areas. 

This paper presents a new course at the intersection of discourse processing and natural language generation (NLG). 
LLMs increasingly perform tasks that require reasoning over lengthy contexts and generating long-form responses, from conversational agents to deep research. Yet, much of what enables these systems to produce meaningful discourse remains poorly understood. 
In addition, recent advances in long-context LLMs have extended the ability of AI systems to process and generate text across thousands of tokens, raising new questions about how coherence, relevance, and discourse structure are maintained over several spans of text. Despite impressive surface fluency, such models often struggle with maintaining global consistency, logical flow, and discourse-level planning. These challenges highlight the need for a deeper understanding of discourse as a fundamental component of language intelligence.

However, courses about (computational) discourse\footnote{Note that ``discourse'' here refers to language processing beyond the sentence boundaries \cite{stede2012discourse}, and not critical discourse theory.} are far and between. Existing curricula also depict distinct partial views of discourse depending on the discipline: for CS students, it may look like a collection of tasks to be solved like topic segmentation, discourse parsing, coreference resolution, and coherence modeling \cite{jurafskyspeech,eisenstein2019introduction}, while for Linguistics students it could be taking concepts from their theoretical classes and understanding their interaction with computational tools.
These views, while valid and useful, do not align well with the reality of today, where multi-sentential processing and generation implicitly but inevitably involve discourse-sensitive capabilities that are not always squarely covered by well-defined tasks.

With these considerations, we designed a new upper-level undergraduate course ``\textbf{Computational Discourse and Natural Language Generation}'', cross-listed between the Linguistics and Computer Science departments. This course was offered at the University of Texas at Austin (UT Austin) in Fall 2025. The course development team spans researchers with diverse background and training: we have three professors with expertise in discourse processing, psycholinguistics, and NLG and its evaluation; a postdoc that has led the development of state-of-the-art language models \cite{lambert2024tulu}; and a computational linguistics PhD student working on discourse and pragmatics.

We collaboratively designed the course content with the following philosophy:
\textbf{(1)} A deep integration of discourse and NLG enables us to connect linguistically-motivated frameworks or theories (e.g., local and global coherence, discourse structure, entity tracking, questions under discussion) with the evaluation, analyses, and improvement of LLMs. This will further teach students how to come up with hypotheses and design controlled experiments for them.
\textbf{(2)} Some class sessions are reserved for mini workshops, helping students to further ground abstract concepts in data.
\textbf{(3)} It is important to foster an explorer's mindset and to engage students in critical thinking, cutting-edge research, and open-ended problems.
\textbf{(4)} Lastly, course content should be accessible but also adequately challenging for students across distinct backgrounds. 
By the end of the course, students will have a richer understanding of how discourse-level linguistic insights can inform computational models, and how advances in NLP can, in turn, provide new insights into the nature of language and communication more broadly.

This paper presents our inspirations (Section~\ref{sec:motivation}), design principles (Section~\ref{sec:design}), course content (Section~\ref{sec:content}), and assignments/projects (Section~\ref{sec:assignments}). We also engaged with a third-party education evaluation (TACC Education Services) to study student reception in the second half of the semester, and these results are discussed in Section~\ref{sec:eval}. Overall, we are optimistic about the timeliness and relevance of a course like this that tightly integrates theoretical and empirical insights, and we conclude with challenges for future iterations to address (Section~\ref{sec:future}).

\section{Motivation and Inspirations}\label{sec:motivation}

To the best of our knowledge, prior or existing courses of a similar kind have treated discourse processing and NLG largely as separate areas.

On the discourse side, we take inspirations from prior graduate courses, such as \textit{Computational Discourse Modeling} offered by Amir Zeldes at Georgetown University, and \textit{Computational Models of Discourse} offered by Alexis Palmer at The University of Colorado Boulder. They both emphasize theoretical and computational approaches to discourse structure and its applications through a graduate-level, project-based seminar.
We also studied individual lectures in NLP courses about discourse, such as Julia Hockenmaier's lectures on discourse coherence and centering theory at the University of Illinois Urbana-Champaign.

In the context of NLG, we consulted the latest openly available courses. We were particularly inspired by Ehud Reiter's lectures on NLG,\!\footnote{\url{https://www.abdn.ac.uk/registry/courses/postgraduate/2024-2025/computing_science/cs551h}} Silvia Casola's lecture on NLG evaluation at LMU Munich,\!\footnote{\url{https://slvcsl.github.io}} Greg Durrett's NLP course offered at UT Austin \cite{durrett-etal-2021-contemporary}, Yejin Choi's NLP course offered at the University of Washington,\!\footnote{\url{https://safe-fernleaf-26d.notion.site/Winter-24-CSE-447-517-Natural-Language-Processing-4142333a001143d2be5ecff1a535c4ab}}
and Song Han's \textit{TinyML and Efficient Deep Learning Computing} offered at MIT.\!\footnote{\url{https://hanlab.mit.edu/courses/2024-fall-65940}}

While these courses equipped students with deep, focused expertise in their respective areas, they also highlighted the need for an integrated treatment that unites discourse and generative models, motivating the design of the current course, ``\textbf{Computational Discourse and Natural Language Generation}''.

\section{Design Principles} \label{sec:design}

Our curriculum is guided by a set of design principles intended to support students in developing a well-rounded and hands-on understanding of discourse processing, modeling, and NLG. These principles highlight the interplay between conceptual frameworks, state-of-the-art techniques, and analytical methodologies, 
with an inclusive pedagogy that takes into account the diversity in students' academic backgrounds and training.

\paragraph{Threading Discourse within NLG.} 
A central principle of the course is to foreground discourse phenomena across various aspects of NLG. Rather than treating discourse as a stand-alone module, we embed core concepts such as coherence \cite{grosz-etal-1995-centering}, discourse reference \cite{karttunen-1969-discourse-referents}, anaphora \cite{beaver2004optimization}, questions under discussion \cite{roberts1996information}, and discourse structure \cite{mann1988rhetorical,webber-etal-2003-anaphora} into modeling and decoding techniques for NLG, as well as the evaluation of LLM outputs. This `threading' approach allows students to see discourse as foundational building blocks for coherent and meaningful long-form outputs from language models, which further helps students think about how information is selected, structured, and realized to account for discourse-level phenomena, instead of just sentence-level correctness.

\paragraph{Mini Workshops for Concept Grounding.} 
We believe it is important for students to critically investigate different viewpoints and engage in contemporary debates on LLMs' capabilities and limitations. This was achieved through two means: (1) A series of in-class group workshops where students apply analytical frameworks and perform annotations on real data. (2) Open-ended, reasoning-based questions with boilerplate code in homework assignments, allowing students to think about computational models, algorithms, and their consequences.
Hands-on work also creates natural opportunities to conduct error analysis and qualitative analyses that one would include in a research paper, for example.

\paragraph{Exposure to Research.} 
We believe in engaging undergraduate students in active research, not only through individual research experiences, but also \emph{in the classroom}.
The fast-paced nature of today's NLP landscape makes it especially important to do so, and we believe that exposing students to recent work that is core to discourse processing, yet not necessarily labeled so, is important in helping them recognize the interplay between theory and practice. These topics are either incorporated into course material organically, or delivered as invited talks from speakers throughout the semester.
This also helps students to develop an appreciation for rigorous dataset and experimental design in research.

\paragraph{Exploration, Not Just Problem Solving.} 
All assignments are intentionally open-ended, encouraging students to define the scope and direction of their work within a given instruction. 
This is our response to the use of Generative AI in education: the focus of assignments for upper-level undergraduate classes should incorporate high-level exploration and project design, rather than problem solving only, which is common in traditional problem sets or exams.
Our structure mirrors real industry and research practice: students must select datasets, justify methodological choices, and articulate the discourse-level phenomenon they aim to improve or analyze. Such design allows students at different experience levels to contribute meaningfully while also fostering creativity and critical thinking.

\paragraph{Accessibility for Diverse Backgrounds.} 
Recognizing that students may come from linguistics, computer science, information science, and other related fields, all content is designed to be accessible without extensive assumptions about prior technical knowledge. Linguistic concepts are introduced with computational relevance, highly technical material such as reinforcement learning is delivered at a level appropriate for the class goal, and programming tasks are scaffolded to accommodate varying levels of experience. 
We also demo common libraries for LLM inference and evaluation in class. 
This inclusive design supports interdisciplinary engagement and positions language-centric perspectives as equal partners to technical ones.

\paragraph{Collaborative Design of Content.} 
A key feature of the course is that it is developed collaboratively by instructors and researchers from multiple institutions and disciplinary backgrounds in linguistics, cognitive science, and computer science. This distributed design process brings together diverse pedagogical traditions and research perspectives, enriching the course content and ensuring that it reflects the multifaceted nature of the `threading' approach mentioned above. Cross-institutional collaboration also allows the curriculum to remain adaptable and responsive to evolving research, teaching practices, and student needs.

\section{Course Content} \label{sec:content}
Below, we provide an overview of the various aspects covered in this course, organized into themes. Note that this does \emph{not} reflect the order of content delivery; as discussed in Section~\ref{sec:design}, we took a threaded approach across all themes. The delivery of the content was mostly built into the lectures with required and optional readings.

\subsection{Content on LLMs and Generation}

\paragraph{Language Modeling and NLG.}
The curriculum starts with an overview of autoregressive language modeling. The goal of this component is to solidify first principles behind language models, and help students understand how they work without a deep dive into specific techniques that improve training. We start with the next word prediction task, moving into transformer-based decoders and training LLM at scale (pre-training, instruction-tuning, and post-training). Decoding algorithms (including greedy, nucleus sampling, \citealt{holtzmancurious}, min-p, \citealt{minhturning}) as well as the basics of long-context models \cite{touvron2023llama,su2024roformer} are also covered.

After this, we provide an overview of NLG by introducing students to NLG tasks from conditional generation such as summarization and translation, to more open-ended ones such as story generation, 
emphasizing how task open-endedness affects both modeling and evaluation. 

Finally, to help students gain a deeper understanding of the inner representations of language models and their multimodal counterparts, we provided them a high-level overview of mechanistic interpretability, covering example papers that used causal mediation, activation patching \cite{golovanevsky-etal-2025-vlms}, early decoding, and activation steering \cite{golovanevsky-etal-2025-pixels}.

\paragraph{RL Overview and Applications.}
Post-training describes different training processes applied to LMs to make them more usable, better instruction followers, and more aligned with human values \cite{liprefpalette}. Common post-training methods employ reinforcement learning (RL), such as RLHF \cite{ouyang2022training} and RLVR \cite{lambert-2011-reperer}. Post-training and RL can also be beneficial in training an LLM to be better at discourse-related tasks. \citet{stiennon2020learning}, for example, use RL with human feedback to improve summarization, and similarly, RL can be used for factuality. Specifically, we give students an introduction on modern post-training techniques and RL for NLP, and then discuss recent research on using RL for discourse-related tasks such as factuality \cite{roit-etal-2023-factually}, question generation \cite{pyatkin-etal-2023-clarifydelphi}, and summarization \cite{stiennon2020learning}.

\paragraph{Demos.} %
The course also included two class sessions devoted to coding demos, focusing on practical implementation and widespread machine learning and LM libraries. 
We designed the demos around giving students the opportunity to practice with the libraries we introduced, and demonstrate how to use documentation on their own.
In the second week of the course we covered basic functionality of \texttt{pytorch} \citep{paszke2019pytorch} and \texttt{transformers} \citep{wolf-etal-2020-transformers}, with a focus on loading LMs, tokenizing input text, and generating text.
In addition, the minimal pair evaluation component was delivered in tandem with a demo involving \texttt{minicons} \citep{misra2022minicons}, a python library that facilitates straightforward, efficient, and fast computation of language model behavioral scores like log-probabilities, and is popularly used to perform minimal pair evaluation.

\subsection{Content on Discourse}

\paragraph{Discourse Structure.} The course covers discourse structure as a key component for analyzing and generating coherent (long-form) text beyond individual sentences, focusing on major linguistically-motivated discourse frameworks or theories such as Rhetorical Structure Theory (RST, \citealt{mann1988rhetorical}), Penn Discourse Treebank (PDTB, \citealt{webber2019penn}), and Questions Under Discussion (QUD, \citealt{roberts1996information}).
Through these frameworks, students learn how to characterize relationships between discourse units: through taxonomies (e.g., \textsc{elaboration}, \textsc{contrast}, \textsc{causal}, and \textsc{temporal}), or through question--answer relations in the case of QUD \cite{wu-etal-2023-qudeval}.
They also learn how these frameworks differ in terms of using tree-like structures (RST and QUD) or being lexically grounded (PDTB); and how QUDs interact with reader expectations \cite{kehler2017evaluating,westera-etal-2020-ted,wu-etal-2024-questions}.
Going beyond, we steer students into thinking about the challenges that lie in modeling these theories and ways to address them. For instance,  discourse relations can be represented as QA pairs, helping to scale data collection \cite{pyatkin-etal-2020-qadiscourse,ko-etal-2022-discourse}. 

We then connect these representations to NLG applications, including: how discourse structure can inform content selection in summarization
\cite{li-etal-2016-role,xu-etal-2020-discourse,liu-zeldes-2023-gumsum,trienes-etal-2025-behavioral} and planning in simplification \cite{wu-etal-2023-elaborative,trienes-etal-2024-infolossqa,liu2025explanatory}.
Overall, this module emphasizes how explicitly incorporating discourse structure bridges linguistic theories and NLG tasks to enable more controlled-generation.

\paragraph{Entity-based Coherence.}
Entities introduced throughout the discourse play a central role in coherence. We introduce the classic Centering Theory \cite{grosz-etal-1995-centering} which tracks the focus of attention, as well as the more flexible Centering in Optimality Theory \cite{beaver2004optimization}. Students work through examples for both accounts in a workshop style, as described in Section~\ref{sec:design}. 
These concepts can be directly used to analyze and assess the coherence of a document. We introduce approaches inspired by Centering Theory, such as Entity Grid \cite{barzilay-lapata-2008-modeling} and DiscoScore \cite{zhao-etal-2023-discoscore}.
Conversely, generation can inform discourse: e.g., summarization informs entity salience \cite{lin-zeldes-2025-gum}.

\paragraph{Entity Tracking.} 
We discuss Entity Tracking, a broad class of problems in discourse understanding that involves representing entities and their states as the discourse unfolds \citep[][\textit{i.a.}]{karttunen-1969-discourse-referents, kamp2010discourse}. Entity tracking has seen a recent reemergence in popularity, primarily as an analytical lens using which researchers can conclude about LLMs' internal representations of the `world' \citep{li-etal-2021-implicit, li2022emergent, kim-schuster-2023-entity}. This content is introduced right after Centering and Entity Grid, to illustrate a rich, multi-facet view of the role of entities in discourse processing.

We specifically used material from \citet{kim-schuster-2023-entity}, a dataset containing textual descriptions of a scenario with entities in a given state followed by a series of state changes. This dataset provides an especially controlled set of stimuli that tease apart shallow processing---e.g., using lexical cues or heuristics that can allow the prediction of entity states without necessarily consulting with the discourse context---from \textit{genuine} entity tracking. 
All in all, the proposed outcome of this component was to: 1) introduce entity tracking as an important discourse-sensitive component of long-context understanding; and 2) understand the relationship between everyday entity-state tracking and classical concepts in theories of coherence.

\subsection{Evaluation}

\paragraph{NLG Evaluation.}
We aim to equip students with both practical knowledge and a critical understanding of NLG evaluation, situating current evaluation practices in a broader perspective, highlighting progress since 2015 alongside persistent challenges such as robustness, replication, impact evaluation, and commercial pressures \cite{ehudreiter2025post}.
We discuss the evaluation of NLG systems, contrasting intrinsic evaluation (which directly measures properties of the generated text, such as fluency or likelihood via metrics like perplexity and entropy) with extrinsic evaluation (which measures downstream application performance). The course comparatively examines reference-based metrics (e.g., BLEU, \citealt{papineni-etal-2002-bleu}, ROUGE, \citealt{lin-2004-rouge}, BLEURT, \citealt{sellam-etal-2020-bleurt}, BERTScore, \citealt{zhangbertscore}) vs.\ reference-free ones such as factuality \cite{tang-etal-2024-minicheck}, LLM-as-a-judge \cite{bavaresco-etal-2025-llms}, and long-context evaluation \cite{liu-etal-2024-lost}.
On a finer-grained level, we discuss how to evaluate LMs' grasp of lexical semantics, such as the discourse particle ``just'' \cite{sheffield-etal-2025-just}.

We also discuss the central role of discourse frameworks in long-form generation tasks: the evaluation of book summaries \cite{changbooookscore}, creative writing \cite{chakrabarty2024art}, discourse diversity \cite{namuduri2025qudsim}, narrative understanding \cite{ahuja2025finding}, multi-turn conversations \cite{laban2025llms}, and AI-story detection \cite{pham2025frankentext}.

\paragraph{Minimal Pair Evaluation.} 
One of our earlier components in this course was to introduce students to the concept of `minimal pair evaluation', a methodology that has been generally used in psycholinguistics-inspired analysis of models \citep{linzen-etal-2016-assessing, warstadt-etal-2020-blimp, misra-etal-2023-comps}. Here, LMs are analyzed as pairs of natural language stimuli that usually differ in a single feature. For instance, for the number-agreement phenomenon, we could compare models on \textit{the keys to the cabinet \textbf{\underline{are}} on the table} vs.~\textit{the keys to the cabinet \textbf{\underline{is}} on the table}. 
Models are typically evaluated based on the probabilities they assign to surface forms. Once we understand their fundamental objective, which is predicting words in context, the concept of minimal pairs emerges as a natural extension for probing their capabilities.
In the context of discourse, the concept most conducive to minimal pair evaluation is that of discourse connectives---words or multi-word expressions that are used to mark discourse relations between arguments given a piece of text. For example, \cref{ex:correct} is more (discourse) coherent than \cref{ex:wrong},\!\footnote{Examples modified from the Winograd Schema Challenge \citep{levesque2012winograd}.} and thus one might hypothesize it to be more likely under a competent LM's distribution. 

\ex. \label{ex:correct} \small \textit{The councilmen refused the demonstrators a permit \underline{\textbf{because}} the \textbf{councilmen} feared violence.}

\ex. \label{ex:wrong} \small \textit{The councilmen refused the demonstrators a permit \underline{\textbf{because}} the \textbf{demonstrators} feared violence.}

We specifically used prior psycholinguistics and LLM-evaluation work on discourse connectives \citep{drenhaus2014incremental, pandia-etal-2021-pragmatic, beyer-etal-2021-incoherence} as our basis for examples and assignments (see Assignment 3 in Section~\ref{sec:assignments}).
Overall, this component introduced the students to methodology imported from the now well-established psycholinguistics toolkit to evaluate LMs \citep{futrell-etal-2019-neural}, and in particular adapt it to test for core components in discourse processing.

\section{Assignments \& Course Project} \label{sec:assignments}

The course involved three assignments, designed to teach students to reason about broad areas that can eventually be incorporated into their course project: basic implementation and analyses of decoding algorithms; understanding the differences between base vs.\ instruction-tuned LMs; discourse connectives, relations, and parsing; local coherence; QUDs; and NLG evaluation. All assignments contain both programming and written portions.

\paragraph{Assignment 1.} 
The first assignment focuses on getting students acclimatized to loading an LM from the \texttt{transformers} library, implementing decoding methods with them, and answering a range of analysis-based open-ended questions. We provided students with basic wrapper code to load an LM, tokenize its input text, and a general decoding function that took as its arguments the input tokens, a few arguments for decoding, and importantly a placeholder logits processing function which implemented the actual decoding algorithm. The students were then asked to implement 1) greedy decoding; 2) top-p sampling \citep{holtzmancurious}; and 3) min-p sampling \citep{minhturning}. 

Our open-ended questions involved asking students to relate top-p and min-p sampling configurations to greedy decoding, and to analyze the resulting outputs from the three algorithms, given a fixed prompt. In the context of story generation, students were asked to analyze both qualitatively and quantitatively, where they were free to implement their own functions for analysis, such as measuring surface form diversity metrics, etc. Lastly, we asked students to compare generations between a standard, next-word prediction LM vs.\ its instruction-tuned counterpart by implementing a perplexity calculation function and computing perplexity of each type of LM on the other's generations. 
Overall, the assignment uses story generation to investigate diversity, and gives students access to toolkits that could lay the groundwork for more complex discourse-sensitive analyses.

\paragraph{Assignment 2.} 
The second assignment focuses on students running three prompting experiments: (1) implementing and comparing BLEU and LLM-as-a-judge \cite{kocmi-federmann-2023-large} for machine translation evaluation; (2) classifying discourse relations (a subset of the DISRPT 2025 shared task datasets, \citealt{disrpt-ws-2025-1}, which cover a wide variety of languages and genres and are represented in a unified format across frameworks) using an LLM-as-a-judge paradigm; and (3) prompting LLMs to generate RST trees. 
These experiments (especially 2 and 3) are open-ended, providing students with the opportunity to practice prompt engineering using the various principles covered in lectures, as well as the opportunity to experiment with a variety of model sizes.
Students were given basic boilerplate code to call models and set up basic formatting, and the data necessary to complete the assignment. Tasks 1 and 2 integrate non-English data, with Task 1 allowing students to bring in their own background by providing their own. Students were also asked to conduct an analysis given various aspects with regard to performance correlation with model size and choice of models, proposing ways to improve the performance, and the errors the model tends to make etc.

\paragraph{Assignment 3.} 
The final assignment touches upon coherence, discourse connectives, and QUDs.
For coherence, students were asked to use Centering Theory and Entity Grid to evaluate the coherence of book summaries generated from \citet{wu2021recursively}.
For the connectives component, we first demonstrated the concept and construction of minimal pairs (echoing our lecture material), and then presented a dataset of minimal pair sentences from \citet{drenhaus2014incremental}, which studied humans' processing of the discourse as modulated by simple changes in discourse connective (thereby adding minimality). We then asked students to load and evaluate an LM of their choice on the dataset, comparing the model's performance with chance-level accuracy, which they had to infer from the dataset, as one of the answers to a question. 
Finally, for QUDs, we asked students to engage with the research question in \citet{wu-etal-2024-questions}: how do QUDs relate to the salience of potential questions? By exploring the generation of potential questions from news articles, judging their salience, and finding out whether they are QUDs, students gain an understanding of expectation-driven QUDs.

\paragraph{Course Project.}
In addition to the three assignments, this course has a final project, letting students showcase what they have learned in a project of their own design. The projects engaged in a variety of topics, e.g., studying new model architecture for generation tasks, unique designs for entity tracking, and investigating multi-lingual applications from a discourse lens. 
These projects are done in groups of $1$-$3$, with several checkpoints during the course to check progress and provide feedback: (1) an initial brainstorming session, (2) a project pitch presentation, (3) a written proposal, (4) a final presentation, and (5) a final written report.

(1) While not a formal requirement, all groups met with either the instructor or the TA to discuss project ideas in the middle of the semester, in advance of the project pitch (2).
This initial phase helped many groups handle project scope, and rule out ideas that were too vague/broad. Formalizing this as a course requirement is encouraged.

(2) Mid-semester, students pitched their projects in an in-class presentation of around six minutes, focusing on motivation, background, and proposed data and methodology. 
Some students also provided pilot results, though we did not explicitly require this.
A portion of their presentation grade came from how well they were able to field questions from the audience, in addition to presentation flow and clarity.

(3) At the end of the semester, students submitted a two-page written proposal for their project.
Again, students were tasked to focus on background, motivation, and proposed data and methodology---this was intended to provide them with groundwork for their final reports.
This also provided students with another chance for more concrete feedback from the instructor and TA.

(4) Students presented their experimental results in the last week of class in a six-minute presentation.
This presentation focused on presenting methodology and (at least some) results, and allowed the instructor, TA, and fellow students to provide additional feedback before the final write-up one week later.
Students were also encouraged to explain any additional experiments they had yet to run for the final report.

(5) The final deliverable is a four-page written report.
We encouraged students to follow the design of an ACL-style short paper, and encouraged them to use their proposal for background, motivation, and related work sections.

\subsection{Logistics + Grading}
All assignments were submitted online in the form of interactive python (e.g., jupyter) notebooks; this enabled students to write the intertwined code and analysis that many of our problems asked for.

Students were given at least two weeks from assignment release to submission.
Since the course took place over $15$ weeks, this allowed for pacing between assignments. 
The last assignment was due two-thirds of the way through the semester, allowing students to focus on the course project towards the end of the semester.
In addition to office hours and in-class time for questions on homework assignments, students had access to Chatter message boards, monitored by the TA, where they could receive help from peers. 
We also encouraged the use of code assistants, provided students included explanatory comments. 
We found this to be quite reasonable, especially since the bulk of assignment questions did not focus on strict implementation, but were instead formulated as mini-experiments.

Given the open-ended design of the assignments, most grading is performed manually.
There is no way around this, as most problems include high-level analysis questions, though a few have a rigid answer key.
For high-level analysis questions, student grades were broken down into three main parts: \emph{code} -- did their code run without errors, \emph{soundness} -- did their code take a reasonable approach to answer the question, and \emph{analysis} -- did they provide conclusions and motivations for those conclusions using results from their code.
These high-level questions either asked for specific points of analysis to be met or were open-ended, in which case they were graded relative to a baseline ``good'' score, with exceptional answers receiving additional points.
Points were allocated primarily to analysis and then soundness, with code usually only being allocated a small portion of the points.

Additionally, since later assignments allow students to use large, closed-source models in addition to smaller, open-source ones, re-running all student code with all models they used would be infeasible in terms of both time and resources.
Instead, for Assignments 2 and 3, student code was verified using a small model as a test-case.
For both open-source and proprietary models, students reported prompts and outputs for grading.

\section{Course Evaluation} \label{sec:eval}

To get insights into student reception and learning experience, we worked with the Texas Advanced Computing Center (TACC) Education Services\footnote{\url{https://tacc.utexas.edu/use-tacc/evaluation-services/}} which conducted an anonymous survey among the students as formative evaluation of this course. Of the $26$ students enrolled in this course at the time of the survey, $19$ responded. The student body consisted of $24$ undergraduate students and two PhD students.

Students enroll in a range of majors (UT Austin allows multiple majors): Computer Science, Linguistics, Mathematics, Statistics and Data Science, Informatics, Robotics, and other Liberal Arts majors. %
$75$\% of the students have prior knowledge of Machine Learning or AI; and $55$\% of the students have prior knowledge in Linguistics. All students have used LLMs prior to taking this course, $75$\% had some experience with deep learning software packages, and $70$\% have used cloud computing/data resources.

\begin{figure}[t]
    \centering
    \includegraphics[width=0.95\linewidth]{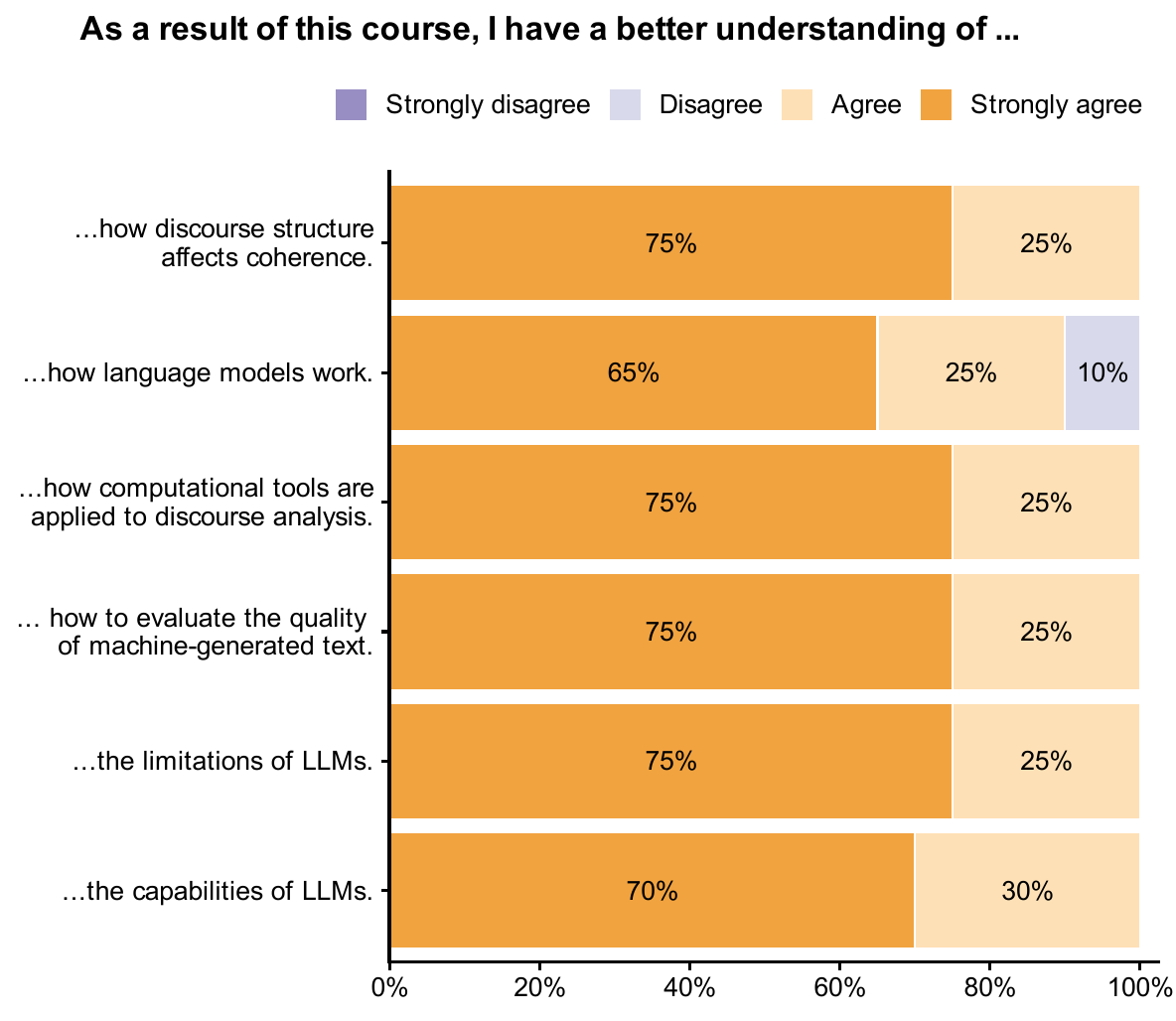}
    \caption{Perceived learning outcomes by topic.}
    \label{fig:survey-result}
\end{figure}

\begin{figure}[t]
    \centering
    \includegraphics[width=0.85\linewidth]{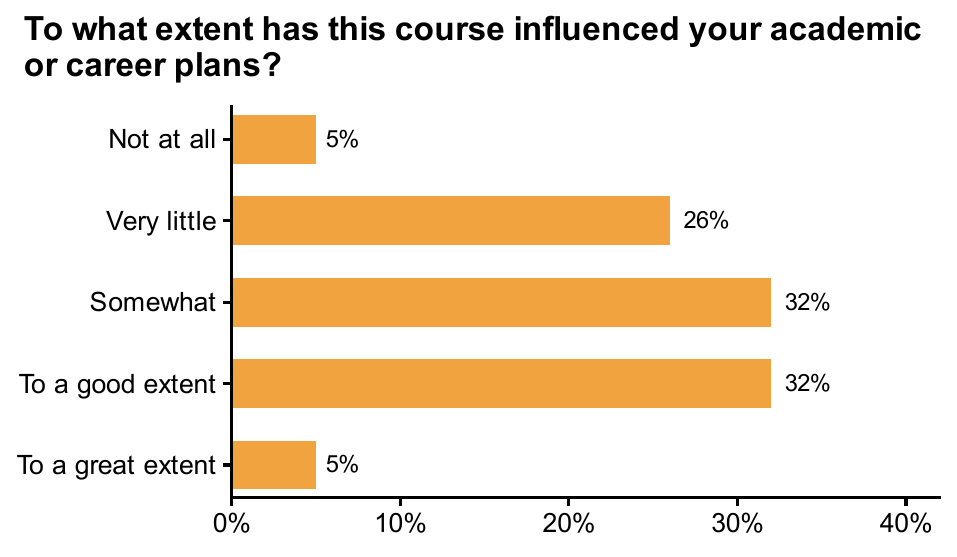}
    \caption{Academic/career impact.}
    \label{fig:survey-result-impact}
\end{figure}

In terms of content difficulty, the majority of the students feel that the course is about the same or more challenging than their prior relevant coursework, with a $95$\% course satisfaction rate. To our delight, $80$\% of the students intend to apply what they learned in this course to their future courses, research, and/or career, and $75$\% said that this course made them want to explore new career paths or fields. In Figure~\ref{fig:survey-result}, we show a topic-wise breakdown of perceived learning outcomes.
Overwhelmingly, students found the assignments and projects valuable to their critical thinking skills and in applying theoretical concepts. 
In Figure~\ref{fig:survey-result-impact}, we show that the majority of students rated this course as having a moderate to significant impact on their academic or career plans.

\section{Challenges and Future Directions} \label{sec:future}

By designing and offering this class, engaging with students, and assessing what they have learned, we are confident that our main course objectives were achieved: to enable students appreciate the deep entanglement of discourse and long-form generation; to perform critical evaluation of theories and empirical findings; and to creatively analyze the capabilities and limitations of LLMs. We believe that these are important skills for the future workforce and our students recognized them as such.

One main challenge for this class is compute resources.
For this first time teaching the course, we were unable to provide students compute resources beyond Colab Pro for Education.\!\footnote{\url{https://colab.research.google.com/signup}}
While this was ultimately sufficient for most students, some assignments and projects would benefit from running models that are too large for this setup.
Providing students with compute resources and/or API credits would be very helpful for this course.

Although most students used LLMs before taking the class, many were concerned about the time it took to run the experiments in the assignments, as well as the time it took for prompt-engineering. We believe this class is a great opportunity to expose them to this, but it is important to discuss these factors in the first lecture in future iterations.

In addition, our homework assignments were designed to be open-ended. This worked well and was positively received by students. However, such open-ended homework assignments did entail manual grading, which is difficult to scale up to a larger class. Innovative ways that combine the spirit of exploration in assignments and automatic grading are needed for scaling this to a larger class size.

Another challenge for this course was the varying levels of student experience with different course topics. Given its interdisciplinary nature, students were from several different majors (primarily CS and Linguistics) and so prior exposure to coding and formal linguistics varied. 
While office hours effectively supported individual students, additional recitation sessions may be needed as class size increases.

In terms of content, this first iteration still remains largely English-centric; we plan to expand our content to other languages in the future.
Future iterations of this course could also include additional applications of discourse and NLG, such as implicit event-argument relations across sentence boundaries \cite{roit-etal-2024-explicating}, the evaluation of long-context QA \cite{xu-etal-2022-answer}, long distance dependencies, and RLHF for improved coherence.

\section*{Acknowledgments}

We thank Lisa Garbrecht, Stephanie Baker, and Yiwen Yang at the TACC Education Services for the formative evaluation of this course.

We thank our invited speakers: Kabir Ahuja, David Beaver, Yapei Chang, Stella Li, Jessica Lin, Chau Minh Pham, and William Rudman. 

We also thank those who provided slides for various portions of this course: Anne Beyer, Antoine Bosselut, Silvia Casola, Tuhin Chakrabarty, Yejin Choi, Greg Durret, Tanya Goyal, Song Han, Julia Hockenmaier, Jaehun Jung, Najoung Kim, Philippe Laban, Nathan Lambert, Nelson F.\ Liu, Ramya Namuduri, Ehud Reiter, Paul Roit, Liyan Tang, and Amir Zeldes.
We thank Yan Cong for providing stimuli for Assignment 3.

The development and evaluation of this course was partially supported by NSF CAREER grant IIS-2145479.

\bibliography{custom,anthology-1,anthology-2}

\end{document}